\documentclass[letterpaper, 10 pt, conference]{ieeeconf}  

\IEEEoverridecommandlockouts                              

\usepackage{graphics} 
\usepackage{url}
\usepackage{booktabs}

\usepackage{amsmath} 
\usepackage{amsfonts}
\usepackage{amssymb}  
\usepackage{multicol}
\usepackage{graphicx}
\usepackage{layouts}
\usepackage{wrapfig}
\usepackage{caption}
\usepackage{graphicx,subcaption}
\usepackage{easyReview}
\usepackage{printlen}
\usepackage{bm}
\usepackage{siunitx}
\usepackage{colortbl}	
\usepackage[utf8]{inputenc}
\usepackage{pgf}
\usepackage{tikz}
\usepackage{tikzscale}
\usepackage{pgfplots}
\DeclareUnicodeCharacter{2212}{\textendash}
\pgfplotsset{compat=1.14}
\usepgfplotslibrary{external}
\usepgfplotslibrary[external]
\tikzset{external/force remake}
\usepackage{algorithm}
\usepackage{algpseudocode}
\usepackage{makecell}
\usepackage{hyperref}

\newtheorem{remark}{\textbf{Remark}}

\newcommand{\state}{\boldsymbol{s}}
\newcommand{\action}{\boldsymbol{a}}
\newcommand{\inputs}{\boldsymbol{x}}
\newcommand{\outputs}{\boldsymbol{y}}
\newcommand{\contact}{\boldsymbol{c}}

\DeclareMathOperator*{\argmax}{arg\,max}

\title{\LARGE \bf
Diffusion-based learning of contact plans for agile locomotion}

\author{Victor Dhédin$^{1*}$, Adithya Kumar Chinnakkonda Ravi$^{1*}$, Armand Jordana$^{2}$, Huaijiang Zhu$^{2}$, Avadesh Meduri$^{2}$,\\ Ludovic Righetti$^{2}$, Bernhard Schölkopf$^{3}$, Majid Khadiv$^{1}$ 
\thanks{$^{*}$The authors contributed equally.}
\thanks{$^{1}$Munich Institute of Robotics and Machine Intelligence, Technical
University of Munich, Germany {\tt\small adithyakumarcr@hotmail.com, \{victor.dhedin, majid.khadiv\}@tum.de}}
\thanks{$^{2}$Tandon  School  of  Engineering,  New  York  University,  USA {\tt\small \{aj2988, hzhu, am9789, ludovic.righetti\}@nyu.edu}}
\thanks{$^{3}$Max-Planck Institute  for  Intelligent  Systems,  T\"ubingen,  Germany {\tt\small bs@tuebingen.mpg.de}}
\thanks{This work was supported by the Max-Planck Institute for Intelligent Systems, New York University and the National Science Foundation grants 1932187, 2026479, 2222815 and 2315396.}%
}

\begin{document}

\maketitle
\thispagestyle{empty}
\pagestyle{empty}


\begin{abstract}
Legged robots have become capable of performing highly dynamic maneuvers in the past few years. However, agile locomotion in highly constrained environments such as stepping stones is still a challenge. In this paper, we propose a combination of model-based control, search, and learning to design efficient control policies for agile locomotion on stepping stones. In our framework, we use nonlinear model predictive control (NMPC) to generate whole-body motions for a given contact plan. To efficiently search for an optimal contact plan, we propose to use Monte Carlo tree search (MCTS). While the combination of MCTS and NMPC can quickly find a feasible plan for a given environment (a few seconds), it is not yet suitable to be used as a reactive policy. Hence, we generate a dataset for optimal goal-conditioned policy for a given scene and learn it through supervised learning. In particular, we leverage the power of diffusion models in handling multi-modality in the dataset. We test our proposed framework on a scenario where our quadruped robot Solo12 successfully jumps to different goals in a highly constrained environment (\href{https://youtu.be/qvIvTUFh_q4}{video}).

\end{abstract}

\section{INTRODUCTION}
Controlling legged robots is a challenging problem, in particular, due to the need to decide over both continuous (e.g., contact forces) and discrete (e.g., which surface patch to step onto next) decision variables. While there has been much progress in designing controllers based on model predictive control (MPC) to tackle the continuous part efficiently~\cite{sleiman2021unified,daneshmand2021variable,meduri2022biconmp,mastalli2022agile}, solving the mixed problem is still computationally intractable. The main goal of this paper is to propose an efficient framework based on a combination of nonlinear MPC (NMPC), Monte Carlo tree search (MCTS), and supervised learning to find feedback policies that decide over discrete variables of the gait.

Several works have tried to solve the generation of whole-body motion through a holistic approach, e.g., using differential dynamic programming (DDP)~\cite{tassa2012synthesis}, contact-invariant optimization (CIO)~\cite{mordatch2012discovery}, contact-implicit trajectory optimization~\cite{posa2014direct}, and phase-based gait parameterization~\cite{winkler2018gait}. While these approaches have shown impressive behaviors in simulation for legged robots, they are not suitable for online motion re-planning, mainly due to their large computation load and being sensitive to initialization. 

To enable real-time motion re-planning, most state-of-the-art frameworks decompose the problem into contact planning and trajectory generation. In this setting, the contact planner decides which end-effector goes to which contact patch, i.e., the discrete part of the decisions to be made. The search over these discrete variables is traditionally done through search-based algorithms~\cite{tonneau2018efficient}, or mixed-integer quadratic programming (MIQP)~\cite{deits2014footstep}. While it has been shown that a relaxed version of the MIQP with L1 norm minimization~\cite{tonneau2020sl1m} can be run on a real quadruped robot in real-time~\cite{risbourg2022real}, these approaches do not scale well when the number of discrete variables grows. Furthermore, these approaches do not take into account the robot dynamics. This makes them impractical for automatically generating dynamic locomotion behaviors with flight phases which is the main focus of this paper.

In~\cite{ponton2016convex,aceituno2017simultaneous}, the authors have included a simplified convex version of the centroidal momentum dynamics in the MIQP formulation. While an efficient implementation of the approach can become real-time capable for a small set of discrete decision variables, this approach can quickly explode as the number of available contact patches increases. To speed up the search problem, ~\cite{lin2019efficient} proposes to learn the outcome of contact transitions using centroidal momentum dynamics. While this is a valid approach for fast contact planning, it was never implemented in a physical simulation or real robot.

As an alternative to model-based planning and control, deep reinforcement learning (DRL) has shown impressive results for agile quadrupedal locomotion~\cite{bogdanovic2022model,zhuang2023robot,cheng2023extreme}. However, none of these works have shown locomotion in highly constrained environments like stepping stones. Recently,~\cite{zhang2023learning} demonstrated locomotion on risky terrains like stepping stones. Nevertheless, as mentioned by the authors, even with the design of several non-trivial reward terms they failed to learn these motions from scratch, mainly due to the sparsity of the environment. Hence, they needed to first train a generalist policy that is able to walk on a simple stepping stone and then fine-tune the policy for each new environment. In contrast, as we demonstrate in this paper, by exploiting the structure of the problem and the environment, our approach requires neither heavy reward shaping nor multi-stage training. 

In this paper, we aim to tackle the problem of contact planning for dynamic maneuvers of legged robots on stepping stones. To do so, we propose to use Monte Carlo Tree Search (MCTS) which has recently been shown to scale well and outperform MIQP for dexterous manipulation~\cite{zhu2023efficient} and gait discovery in locomotion~\cite{amatucci2022monte}. In particular, and unlike previous approaches, we use nonlinear model predictive control (NMPC) inside the MCTS formulation to ensure the feasibility of the motion in a closed-loop sense. In contrast to MIQP approaches that resort to a simplified model of the system dynamics~\cite{ponton2016convex,wang2021multi}, we do not have such simplification and roll out the whole-body NMPC to evaluate a contact plan. This way, we can make sure that the contact plan is consistent with the limitations and constraints of the whole control pipeline. This is in contrast with~\cite{tsounis2020deepgait,yu2021visual} which learns the contact planner in isolation of the low-level policy that realizes it. 
In contrast to~\cite{gangapurwala2022rloc,yu2021visual}, we cast the problem as MCTS whose convergence properties are better understood compared to continuous RL. Also, contrary to~\cite{gangapurwala2022rloc} that only find a single solution, we collect all feasible solutions for any goal position and use a diffusion model to encode the multi-modality in the dataset.

We show that through learning the feasible solutions of the MCTS, we can have reactive policies that can select the next contact patches as a function of the current state of the robot and the environment. Through extensive simulation experiments of quadruped jumping on stepping stones, we show the effectiveness of our proposed framework. It is important to note that stepping stones are chosen as an extremely sparse environment, hence our approach is trivially applicable to any environment together with a segmentation algorithm \cite{deits2015computing,grandia2023perceptive}. 

The main contributions of the paper are as follows:
\begin{itemize}
    \item We present a novel framework that takes the constraints of the robot's low-level controller and the environment into account to efficiently control legged robots in highly limited environments (stepping stones),
    \item Through a systematic comparison, we show that diffusion models are an ideal candidate to learn from expert data when multi-modality is present in the dataset,
    \item We demonstrate automatic online surface selection for dynamic quadrupedal locomotion through a learned feedback policy.
\end{itemize}


\begin{figure}
    \centering
    \includegraphics[width=\linewidth]{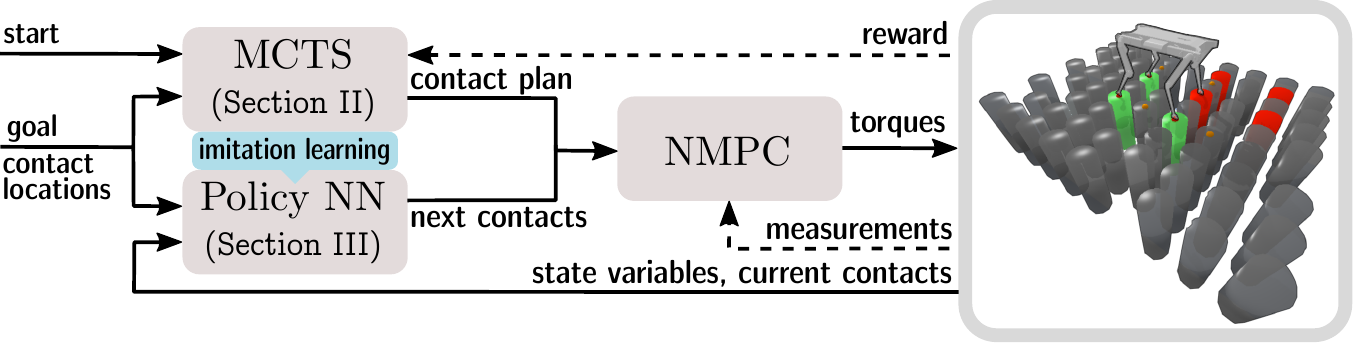}
    \vspace{-5mm}
    \caption{A schematic block diagram of the framework. We use the NMPC formulation in~\cite{meduri2022biconmp}.}
    \label{fig:block}
    \vspace{-6mm}
\end{figure}

\section{Contact Planning using MCTS}\label{sec:MCTS}
We formalize our contact planning problem as a Markov-decision process (MDP), where the state $\state$ represents the location of end-effectors in contact with stepping stones which we denote by a discrete index, and the action $\action$ selects the next locations for each end-effectors and brings the system to a new state $\state ' = f(\state,\action)$. Each state is evaluated by a reward function $r(\state)$ that specifies its associated immediate reward.

To solve this problem, MCTS creates a search tree $\mathcal{T} = (\mathcal{V}. \mathcal{E})$ where the set of nodes $\mathcal{V}$ contains the visited states and the set of edges contains the visited transitions ${(\state \overset{\action}{\rightarrow}\state ')}$. Each transition maintains the state-action value $Q(\state, \action)$ and the number of visits $N(\state, \action)$. MCTS grows this search tree iteratively by the following steps.

\begin{enumerate}
  \item \textbf{Selection}: Start from the root node (initial state) and select successively a child until a leaf (node that has not been expanded yet or terminal state) has been reached. If all the children of a node have already been expanded, a child is selected according to its Upper Confidence Bound (UCB) ~\eqref{eq:ucb} that balances exploration and exploitation during the search.
  \item \textbf{Expansion}: Unless the selected state from the previous step is a terminal state, its successor states are added to the tree by enumerating all possible actions. The corresponding state-action pairs are initialized with $Q(\state, \action)=0$ and $N(\state, \action)=0$.
  \item \textbf{Simulation}: From one of the successor states, random actions are performed onward for a predefined number of steps to create a simulation rollout. The reward $r$ is evaluated at the end of the simulation. 
  \item \textbf{Back-propagation}: The reward is then back-propagated to update the state-action value $Q(\state, \action) = Q(\state, \action) + r$ and the number of visits $N(\state, \action) = N(\state, \action) + 1$ for all the states along the node selected in the selection and expansion steps.
\end{enumerate}

\begin{figure}
    \centering
    \includegraphics[width=0.5\linewidth, angle=0., trim={0.8cm 14.3cm 0.8cm 0.8cm}, clip]{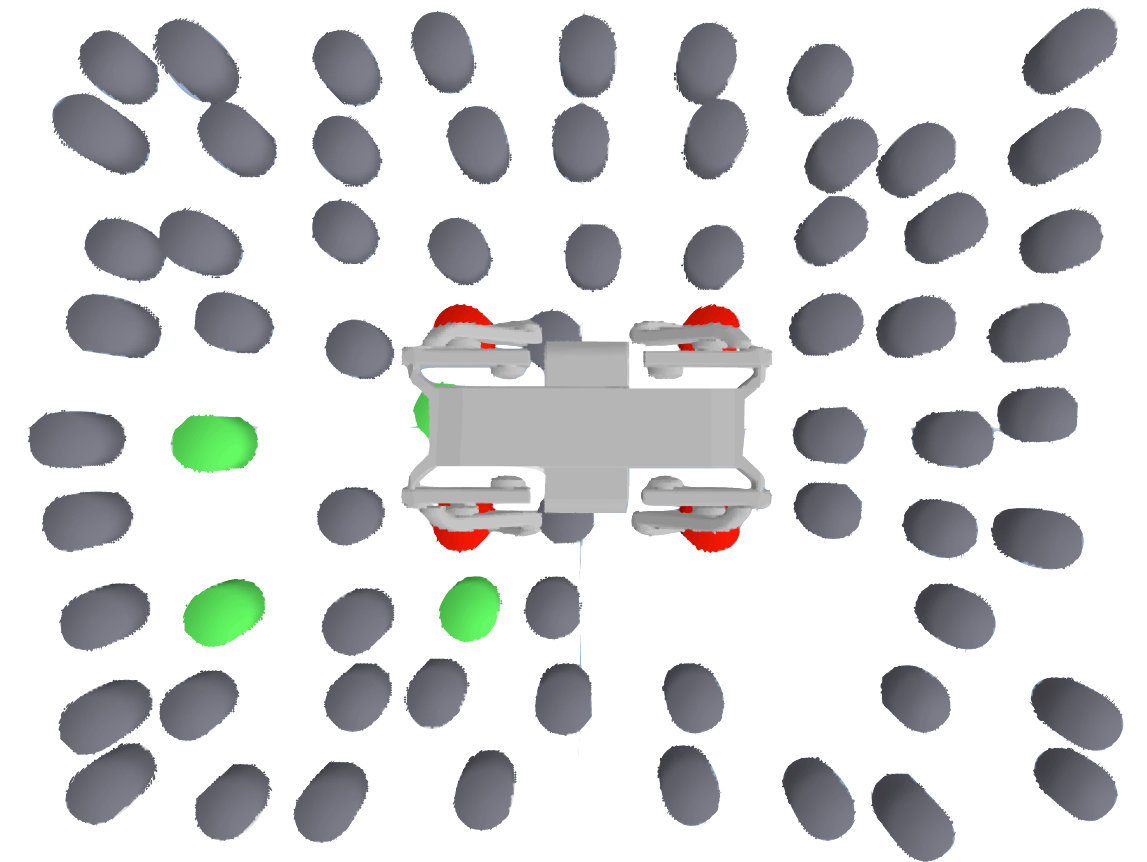}
    \vspace{-1mm}
    \caption{Top view of the environment. The stepping stones in green represent the goal contact locations, and the stepping stones in red represent the start contact locations. Some stepping stones are randomly removed.}
    \label{fig:stageArea}
    \vspace{-7mm}
\end{figure}

As shown schematically in Fig.~\ref{fig:stageArea}, in our problem formulation, the MCTS is given the initial and final desired locations of all the robot end-effectors and is asked to compute the feasible set of patches that results in a successful motion. 
To balance between the exploration of unvisited and visited states, we consider the upper confidence bound (UCB)~\cite{kocsis2006bandit} for each node. In the \textbf{Selection} phase, MCTS selects the action with the highest UCB score $\action = \argmax_{\action} U(\state, \action)$ where the UCB score is defined as
\begin{align}\label{eq:ucb}
U(\state,\action) = \frac{Q(\state,\action) }{N(\state,\action)} + c\sqrt \frac{\log N(\state)}{N(\state,\action)} 
\end{align} 
where $c$ is a coefficient to balance exploration against exploitation and $N(\state) = \sum_{\action} N(\state,\action)$ is the total number of visits for a node. The reward function that is used to update the state-action value is
\begin{align}\label{eq:reward}
    r(\state) = W \sigma(\frac{1}{N_e} \sum_{j=1}^{n_e} (1 - \frac{\vert\vert \mathbf{c}^{j}_{W} - {}^g\mathbf{c}^{j}_{W} \vert\vert_2}{d_{max}}) )
\end{align} 
where $\mathbf{c}^{j}_{W}$ is the contact location in world frame of the $j$th end-effector at state $\state$, ${}^g\mathbf{c}^{j}_{W}$ is the desired goal location for the $j$th end-effector in world frame, and $N_e$ is the total number of end-effectors. $d_{max}$ is the maximum distance between two contact patches in the map. $\sigma : [0,1] \rightarrow [0,1]$ is a function that shapes the reward. $W \in \{-1, 1\}$ is a success indicator, by default set to $1$ when the goal state is not reached.


The following adjustments have been made to improve the efficiency of the algorithm.
During the \textbf{Expansion} phase, to limit the search space, we additionally perform a simplified kinematic feasibility check to prune the sequences that are likely to be not reachable or cause self-collision for the legs of the robot. This check verifies that the size of the step taken by each foot is below a maximal step distance $d_{step}$ and that the legs are not crossing.
It is important to note that, similar to~\cite{labbe2020monte}, since we can define a reward at each state that gives us a proxy of how close this state is to the end goal, we do not simulate a complete rollout to a terminal state as opposed to game-play (such as chess or go). Instead, we only take one action at the chosen state and back-propagate the reward from the resulting next state, hence a rollout depth of one in the \textbf{Simulation} phase. Moreover, we perform whole-body NMPC using the found sequence of contact locations only when a terminal state is reached (all the end-effectors are at the desired goal surface patches). This way, we avoid performing whole-body NMPC for every rollout in the \textbf{Simulation} phase. This significantly reduces search time as whole-body NMPC is computationally intensive. If this contact sequence leads to dynamically feasible whole-body motion physical simulation, $W$ is set to $1$, else to $-1$. Ultimately, we back-propagate the final reward.
Overall, the MCTS searches for promising kinematically feasible contact sequences, and the NMPC evaluates their dynamic feasibility at the end.

We use the NMPC formulation in ~\cite{meduri2022biconmp} to perform the rollouts in MCTS and evaluate kinematically feasible contact patch candidates. Our NMPC is based on a decomposition of the problem into a kinematic and dynamic optimizer. In this setting, given a contact plan (which end-effector goes to which surface patch), the dynamic optimizer takes into account the centroidal momentum dynamics and constructs a finite-horizon optimal control problem to find a feasible set of contact forces and centroidal trajectories. To track these momentum trajectories, while penalizing contact constraints, a whole-body kinematic optimal control problem is solved to output the desired whole-body accelerations. The contact forces from the dynamics optimizer and whole-body joint trajectories from the kinematic optimizer are then used within the full dynamics of the system to compute joint torques. Finally, a joint low-impedance controller is added to the feed-forward torques to account for model errors and uncertainties. The kinematic and dynamic optimizers re-compute trajectories at 20 Hz and the joint torques are computed at 1 kHz and are sent to the robot.

\begin{remark}
    In our NMPC formulation, we keep the contact locations fixed and consider them to be at the center of the stepping stone. The stepping stones considered are small, this simplification is not limiting in terms of the number of feasible motions found. However, we can still optimize for the location of the feet within the stepping stones patches in the NMPC. This might require representing the contact locations in a form other than 3D points. We leave this modification for future work.
\end{remark}

\begin{remark}
    While we can consider all types of gaits and terrain shapes to search over, to reduce the search space, here we only consider jumping and trotting motion over stepping stones. Note however that the framework can be deployed to consider even a general acyclic gait at the cost of extremely larger search space. In that case, we would need to have different cost weights for different gaits. The jumping gait, even if rarely desired in practice, showcase the ability of the planner to plan for highly dynamic motions.
\end{remark}

\section{Learning Contact Planner}\label{sec:learning}

While efficient, our MCTS together with the NMPC framework cannot be run in real-time. To enable the robot to reactively select the next feasible contact patches given the current ones, we learn a neural network to imitate the MCTS policy via supervised learning. While MCTS admits a natural extension of a learnable value function and action probability prior~\cite{silver2017mastering, schrittwieser2020mastering}, we decided not to adopt this methodology despite its success in game-play for two reasons. First, in contrast to generic game-play, locomotion tasks on different maps (e.g. varying locations and numbers of stepping stones) are likely to have different states and action space; an MCTS trained on a specific map does not generalize to other environment maps. Second, some other sensory inputs that are not modeled in the MCTS state space may provide additional information (e.g. base velocity) on \emph{if} and \emph{where} to make the next contact. Therefore, we take an imitation learning perspective and treat the MCTS as an algorithmic demonstrator, whose behavior will be cloned by a neural network policy. 

\begin{figure}
    \centering
    \includegraphics[width=\linewidth]{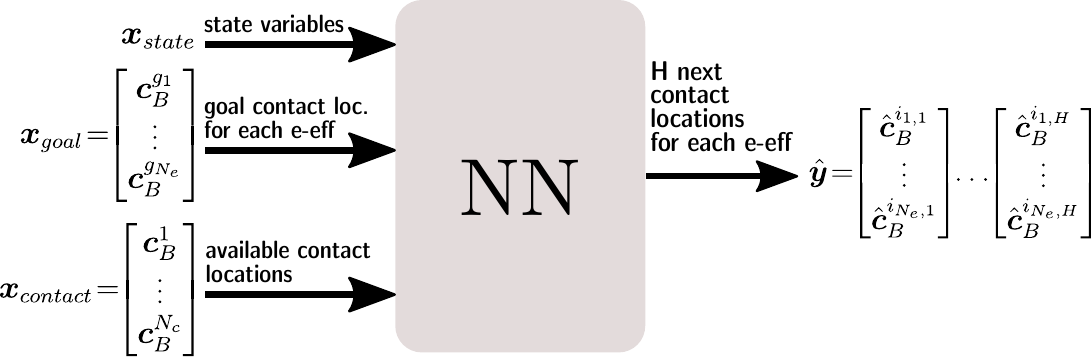}
    \caption{High-level contact planning network policy}
    \label{fig:high_level_network}
    \vspace{-8mm}
\end{figure}

We collect data of the dynamically feasible solutions discovered by the MCTS together with the contact locations of the simulation environment. This dataset is then used to train a neural network in a supervised fashion. A schematic structure of the network is shown in Fig. \ref{fig:high_level_network}.
To be independent of the global position of the robot and environment, we express all positions in the inputs and outputs to the network with respect to the current base frame $B$ of the robot.
The inputs, $\inputs$ to the network are all the current 3D position of the available contact locations $\boldsymbol{x}_{\text{contact}}$, denoted by $\contact_B^i \in \mathcal{C}$ with $i \in \{1, ..., N_c\}$ and $\mathcal{C}$ the set of available contact locations ($N_c$ is the current number of available contact locations), $\boldsymbol{x}_{\text{state}}$ that includes some state variables (see Sec. \ref{subsec:setup}) and $\boldsymbol{x}_{\text{goal}}$, the final desired $N_e$ end-effector locations ($N_e$ is the number of end-effectors). The size of the input is equal to $3 \times (N_{c} + 2 \times (N_{e} + 1))$.

The neural network policy predicts the future $H$ contact locations for each end-effector, denoted by $\hat{\contact}_B^{i_{j,h}}$ for the end-effector $j \in [1, \hdots, N_e]$ and the next step $h \in [1, \hdots, H]$. The size of the output $\hat{\outputs}$ is equal to $3 \times H \times N_{e}$.  As the NMPC is quite sensitive to small errors in the contact locations, we project the contact locations given by the policy to the closest center of the patch. We refer to the projection function as $p_{\mathcal{C}}$ in the following. We show in Sec.~\ref{sec:navigation_results} that the distance between the contact locations given by the policy and the projected ones is way lower than the half distance between the two stepping stones. This means that the choice of contact patch by the policy is not affected by this projection. 

\subsection{Problem specifications}\label{subsec:setup}

In our setup, a quadruped robot ($N_e=4$) navigates in an environment with up to $81$ stepping stones. Each stone provides the robot with a cylindrical patch of radius $4.4 \, \text{cm}$ and height $h = 10 \, \text{cm}$ to step onto. The stepping stones initially form a regular grid of spacing $(e_x, e_y)$ so that the feet lay on $4$ stepping stones in the initial configuration. From this, the environment is randomized. The position of each stone is then displaced by $\epsilon_x (\frac{e_x}{2} - r) $ with $\epsilon_x \sim \mathcal{U}(-\alpha_x, \alpha_x)$, $\alpha_x \in [0,1]$ in the $x$ direction (respectively for the direction $y$). Similarly, the randomized height equals $(1+\epsilon_h)h$ with $ \epsilon_h \sim \mathcal{U}(-\alpha_h, \alpha_h), \alpha_h \in [0,1]$. Additionally, $N_{\text{removed}}$ stones are randomly removed. The simulated environment can be seen in Fig.~\ref{fig:stageArea}.
Goals are also sampled randomly so that the center of the 4 goal contact locations is within $d_{min}^{g}$ and $d_{max}^{g}$ of the initial robot position.

In our experiments, $H$ encodes the NMPC horizon, which in our problem is two jumps in the future ($H = 2$). The goal is to evaluate the learned policy before each jump and feed the selected contact locations to the NMPC. The planner only provides the set of contact locations required by the low-level controller. One could choose a longer horizon, but it is unnecessary as, at runtime, the planner is called every gait cycle and updates the plan changes depending on the current state of the robot. Note that goals can be set at any distance from the start. As state variables, we consider the position of the end-effectors, the current base linear, and angular velocities (all expressed in base frame $B$).

\subsection{Dataset}\label{sec:data_recording}

To collect a diverse dataset, we sample a random environment and run MCTS for a fixed maximum number of iterations. To collect diverse paths towards a goal, we keep up to $N_{\text{paths}}$ different feasible paths for the same goal and environment. 
To cover a wider range of robot states, for each MCTS solution, we perturb the simulation $N_{\text{rand}}$ times and add feasible solutions to the dataset. The randomization procedure consists of randomizing the initial state of the robot (position and velocity) as well as the contact locations inside the selected patches. The training data $(\inputs, \outputs)$ (see Fig. \ref{fig:high_level_network}) are recorded at each jump ($\outputs$ are the contact locations of the next two jumps).  
We repeat the procedure on $N_{\text{env}}$ different environments (set of stepping stones).

\subsection{Neural network architectures}\label{sec:architecture}

Our learning problem structure is a selection procedure as the policy should ideally return contact locations that are given as input. While this can be achieved using a projection function, some network architectures are suited to this task such as the Pointer-Network architecture (\ref{sec:PointerNet}) that we consider as a candidate.
Additionally, our dataset is multi-modal as MCTS provides different contact sequences for the same start and goal contact locations in a given environment. It is not possible to represent such multi-modal data distribution with a conventional uni-modal policy class as the model could collapse to one of the modes or an average over several modes (see Fig \ref{fig:results_diagonal}). Therefore, we consider denoising diffusion probabilistic models (DDPMs)~\cite{ho2020denoising} (cf.~\ref{sec:DDPM}), as another potential candidate, since they are theoretically grounded to handle multi-modality~\cite{block2023provable} and is practically verified for some robotic applications 
\cite{janner2022planning}~\cite{chi2023diffusion}. We also consider multilayer perceptron (MLP) architecture as a baseline.

\subsubsection{Pointer-Networks}\label{sec:PointerNet}

Pointer-Networks~\cite{vinyals2017pointer} take as input a sequence and output discrete indices, called \emph{pointers}, that select elements from the input sequence. In this case, the projection $p_{\mathcal{C}}$ is not performed as the model directly outputs from the input set. The architecture is composed of $2$ recurrent networks and an attention mechanism that operates on the past decoder's hidden states and all the encoder's hidden states.
At each step, the output of the decoder is the index of the encoder's hidden state that has the maximum attention value with the past decoder's hidden state. This operation is repeated as many times as needed, in our case $8$ times (the next two contact locations for each foot).

To make the model select only the contact patches from the input $\inputs$, $\inputs_{\text{contact}}$ is given as the input sequence while $[\inputs_{\text{state}}, \inputs_{\text{goal}}]$ is embedded and given as the first hidden state of the encoder.
Like so, contact patches can be provided in any order and a different number which is not the case for instance for an MLP.

\subsubsection{Diffusion models}\label{sec:DDPM}

DDPMs are generative models that map samples from a latent random distribution to the data distribution in $T$ steps by successive denoising of the original noise.
For each intermediate step $t \in [1, T]$, one can sample a corrupted input $\inputs_t$ by adding noise $\boldsymbol{\epsilon}_t$ to a sample of the data $\inputs_0$. A variance schedule assigns an increasing noise level at each step $t$ so that $\inputs_T$ can be seen as a pure random noise (usually from a Gaussian distribution). Those corrupted samples are used to train the diffusion model $\boldsymbol{\epsilon}_{\boldsymbol{\theta}}$, parametrized by $\boldsymbol{\theta}$, to estimate the noise added in a supervised manner. It is done by minimizing the MSE loss between the actual sampled noise $\boldsymbol{\epsilon}_t$ and the estimated one $\boldsymbol{\epsilon}_\theta(\inputs_0 + \boldsymbol{\epsilon}_t, t)$. Minimizing the MSE loss leads to the minimization of the variational lower bound of the KL-divergence between the data distribution and the distribution of samples drawn from the DDPM~\cite{ho2020denoising}.

To sample with the trained model, noise is successively removed from a random sample $\inputs_T$ in the following way
$$\inputs_{t-1} = a_t \left( \inputs_t - b_t \, \boldsymbol{\epsilon}_\theta(\inputs_t, t) \right) + \sigma_t \boldsymbol{z}$$
where $\boldsymbol{z} \sim \mathcal{N}(\mathbf{0}, \mathbf{I})$, and $a_t$, $b_t$, $\sigma_t$ are computed according to the noise schedule.

\begin{figure}
    \centering
    \includegraphics[width=0.6\linewidth]{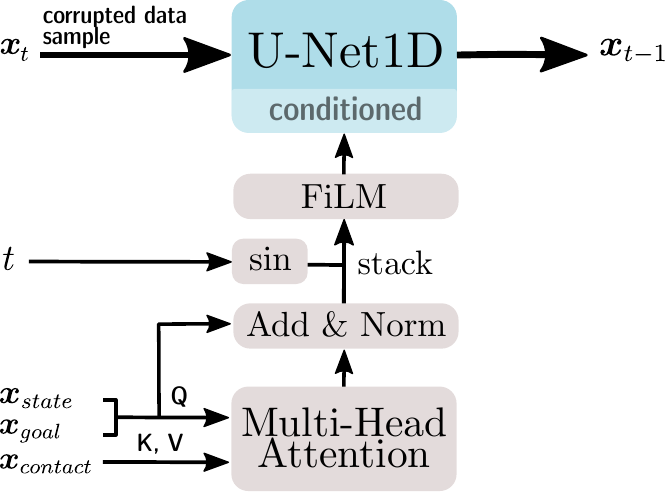}
    \caption{High-level description of the conditional U-Net1D with multi-head attention. \textit{sin} refers to Sinusoidal positional embedding applied to $t$.}
    \label{fig:UNet_architecture}
    \vspace{-7mm}
\end{figure}

U-Net-based architectures are widely used as DDPMs, especially for conditional image generation. Following~\cite{janner2022planning}, we consider a conditional U-Net1D with 1D convolutions applied on the input sequence length (end-effector dimension). The conditioning is done on both the denoising iteration $t$ of the diffusion process and the current input $\inputs$. Similarly to what has been done with the Pointer-Network in Sec.~\ref{sec:PointerNet}, the input $\inputs$ is split into two. We apply a multi-head attention layer~\cite{vaswani2023attention} with $[\inputs_{\text{state}}, \inputs_{\text{goal}}]$ as query and $\boldsymbol{x}_{\text{contact}}$ as key and values. Thus, contact locations can be shuffled and provided in any number. The resulting embedding of $\boldsymbol{x}_{\text{state}}$ is concatenated to a sinusoidal position embedding~\cite{vaswani2023attention} of $t$ to form the input of a feature-wise linear modulation (FiLM)~\cite{perez2017film} layer, as proposed by~\cite{chi2023diffusion}. The architecture is detailed in Fig.~\ref{fig:UNet_architecture}. Finally, the conditioning vector is added to all layers of the encoder and the decoder parts of the network, as done in~\cite{janner2022planning}.

\section{Results and Discussion}\label{sec:results}

In this section, we present the results of applying the proposed framework to control the Solo12 quadruped robot~\cite{grimminger2020open} on stepping stones in the Pybullet simulation environment~\cite{coumans2016pybullet}. We start by showing the capability of the MCTS together with NMPC to physically realizable contact plans on stepping stones (\ref{sec:mcts_results}). We also show that our method can be extended to a trotting gait (\ref{sec:trotting_gait}). In addition, to emphasize that the contact plans found by MCTS are non-trivial, we compare our approach with a naive contact planner (\ref{sec:naive_planner}). About the learned policies, we first compare the capability of different neural network architectures to handle multi-modality in a simple example scenario (\ref{sec:preliminary_results}).
Then, we present the result of using the diffusion model to learn a policy that can output contact plans from the state of the robot and the environment (\ref{sec:navigation_results}).
Finally, we show the effectiveness of our learned policy in selecting the feasible contact patches as the environment and goal change dynamically.

\subsection{MCTS contact planner}\label{sec:mcts_results}

To test the effectiveness of the MCTS with NMPC in generating feasible jumping motions on stepping stones, we run the algorithm for $10^4$ iterations in highly randomized environment: $N_{\text{removed}}=9$, $d_{min}^g = 28$cm, $d_{max}^g = 42$cm, $\alpha_x = \alpha_y = 0.9$ and $\alpha_h = 0.25$.  For each episode, there are $N_{\text{box}} = 81 - 9 = 72$ stepping stones in the environment. Given that for jumping motion all $N_{e}$ end-effectors are in contact, the number of configurations for each jump is equal to $N_{\text{box}}^{N_{e}}$. This rules out the use of a brute-force search to solve the problem.
Pruning drastically reduces the number of possible configurations, decreasing our branching factor to 324 at maximum, which is still high.

As hyperparameters, we found out that $c = 0.01$ and $\sigma = sigmoid(T(x-1))$ with $T=5$ leads to a sufficient balance between exploration and exploitation. $\sigma$ shapes the reward so that it is significantly increased for a state close to the goal state leading to a better exploitation for nodes close to the terminal node. We chose $d_{step} = 24$ cm to prune long jumps that are likely to fail.

\begin{figure}
    \centering
    \includegraphics[width=0.9\linewidth]{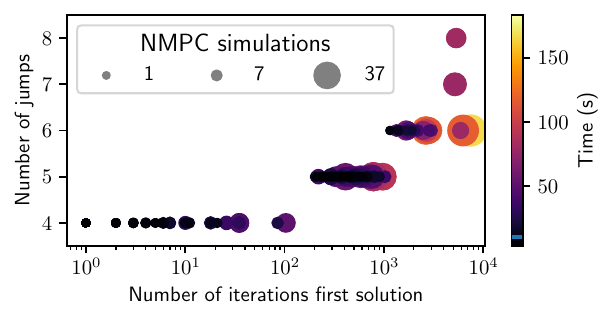}
    \vspace{-3mm}
    \caption{performance analysis of the MCTS. The median search time is 9.65s (blue line on the color bar). After 5000 iterations, it is unlikely to find a feasible contact plan (here only 6 out of 500).}
    \label{fig:MCTS_perf}
    \vspace{-4mm}
\end{figure}

We evaluate our method over $500$ environments by estimating the search time, the number of iterations, and the number of NMPC simulations performed to find the first feasible solutions with a maximum number of iterations equal to $10^4$. MCTS was able to find $410$ feasible contact plans (success rate of $82 \%$). For those, the median search time is $9.65$ sec, the median number of iterations is $35$ and the average number of NMPC simulations performed is $5.9$. The search time depends mostly on the contact sequence length and the number of NMPC simulations, as shown in Fig.~\ref{fig:MCTS_perf}. On average, the NMPC simulations take about 81$\%$ of the total planning time to find the first solution. Indeed, by doing the same experiment without removing any stepping stones, MCTS found $94$ solutions out of 100 under $10^4$ iterations.

\begin{figure}
  \centering
  \begin{subfigure}{0.32\linewidth}
    \centering
    \includegraphics[width=\linewidth]{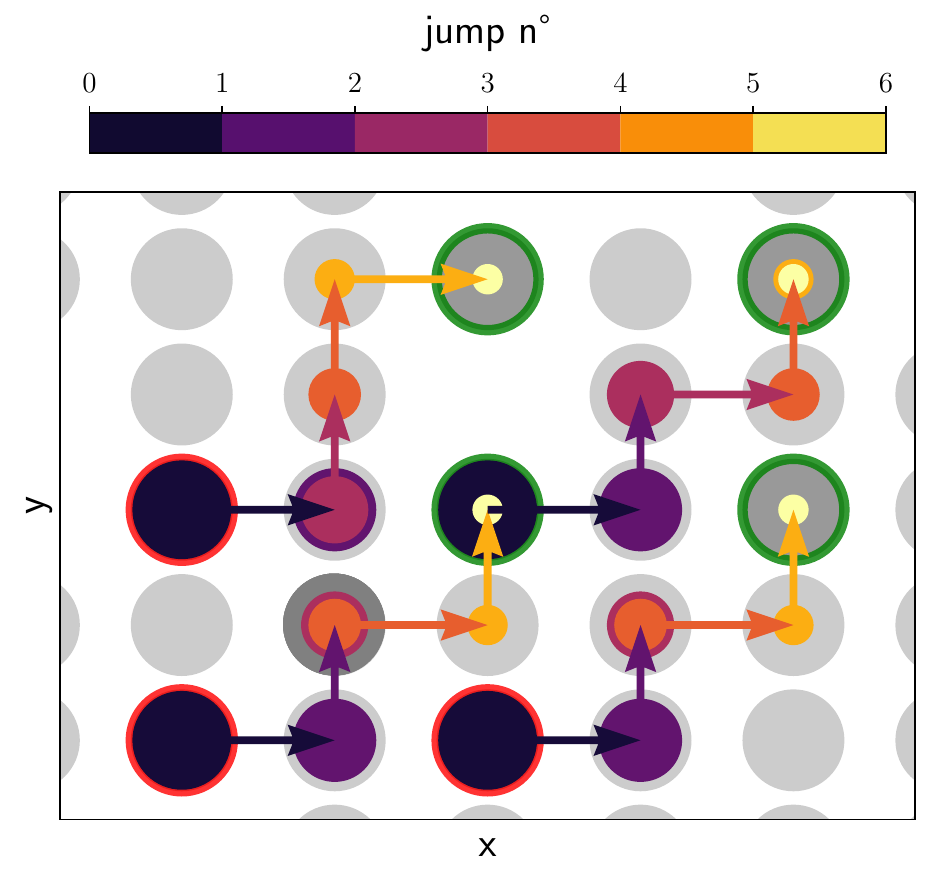}
    \vspace{-1.cm}  
    \caption*{}
    \label{fig:MCTSseq0}
  \end{subfigure}
  \begin{subfigure}{0.32\linewidth}
    \centering
    \includegraphics[width=\linewidth]{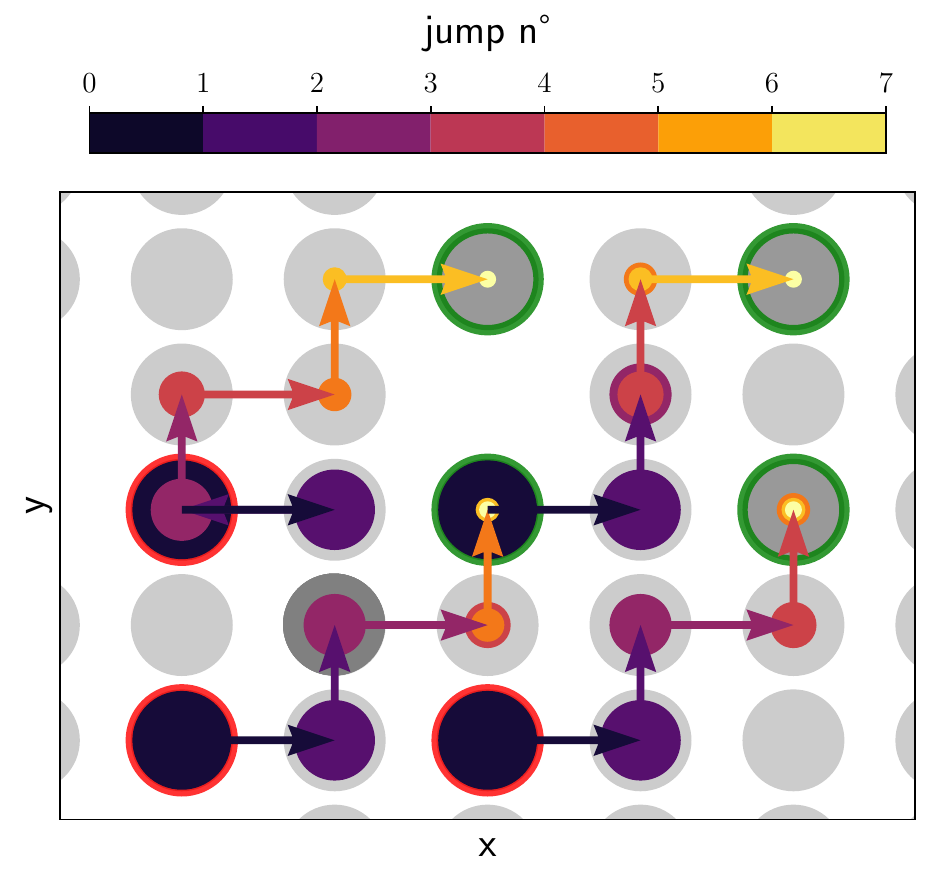}
    \vspace{-1.cm}  
    \caption*{}
    \label{fig:MCTSseq1}
  \end{subfigure}
  \begin{subfigure}{0.32\linewidth}
    \centering
    \includegraphics[width=\linewidth]{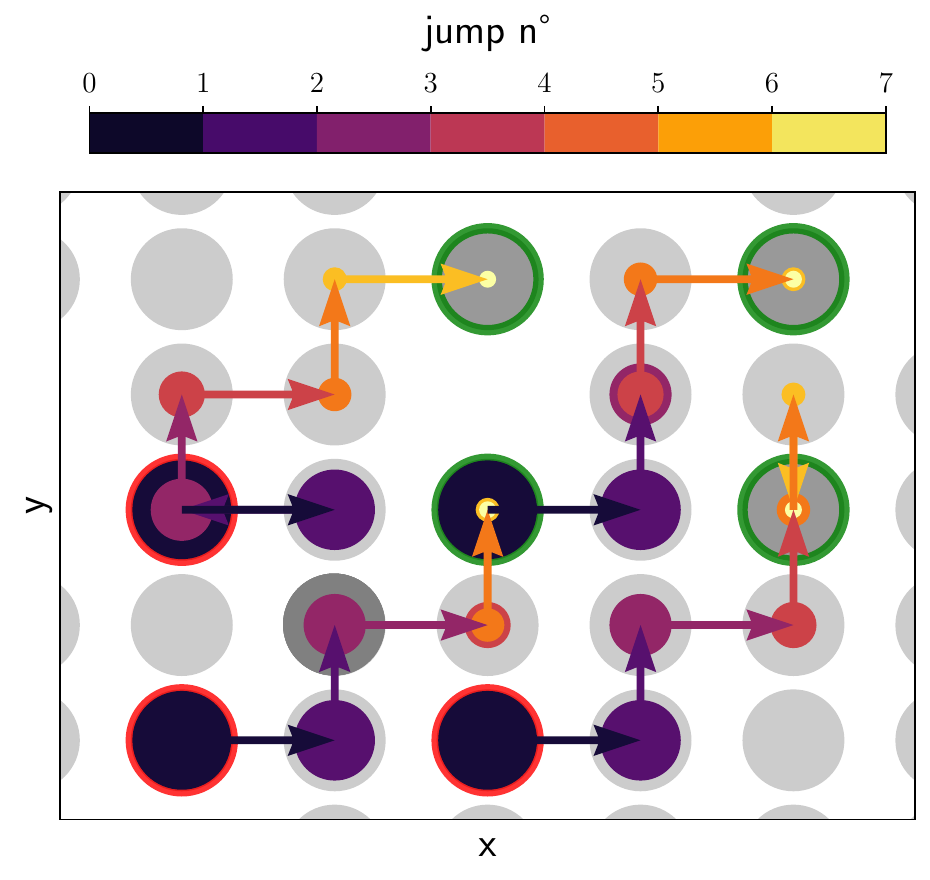}
    \vspace{-1.cm}  
    \caption*{}
    \label{fig:MCTSseq2}
  \end{subfigure}
  \caption{Three contact sequences generated by MCTS on the same environment. Each path following the arrows is a contact sequence for an individual foot. Starting contact locations are circled in red and goal contact locations in green (the contact location at the center circled in green is both a starting and goal contact location).}
  \label{fig:mcts_contact_plans}
  \vspace{-5mm}
\end{figure}

Qualitatively, contact plans found for the same goal are quite similar to each other as MCTS explores nodes close to the goal more often, as they have a higher average value and have been visited less. Multi-modality is more pronounced at short horizons, as can be seen in Fig.~\ref{fig:mcts_contact_plans}. Some interesting and non-trivial motions found by the MCTS are also shown in the accompanying \href{https://youtu.be/qvIvTUFh_q4}{video}.

\subsection{Extension to trotting gait}\label{sec:trotting_gait}

To demonstrate that our framework can generalize to other gaits, we show that different physically feasible contact plans can also be found by only modifying the gait parameters and costs of the NMPC. We experimented with a slow trot gait in the following (gait period of $0.9$ seconds, stance percentage of $84\%$). The parameters of the gait are fixed during the search.

We performed the search on $300$ new environments, randomized in the same way as for the jumping gait. MCTS successfully found feasible contact plans in $68\%$ of the cases. The search is slower compared to the jumping gait, with a median search time of $12.3$ seconds. This could be explained as trotting in this case is more demanding, as it constrains more the possible motions of the legs. The quadruped stands on two opposite legs while reaching the next contact locations, which often cause imbalance. Qualitative results of the trotting gait are shown in the accompanying \href{https://youtu.be/qvIvTUFh_q4}{video}.

\subsection{Comparison with a naive planner}\label{sec:naive_planner}

To show the necessity of using MCTS in such challenging environments, we implemented a naive contact planner based on the Raibert heuristics used in \cite{kim2019highlydynamicquadrupedlocomotion} which gives the footstep locations to follow in order to achieve the desired velocity.
Our naive contact planner policy projects the output of the Raibert heuristics to the closest stepping stones of the environment (similar to \cite{grandia2023perceptive}). The desired velocity is set in the direction of the goal and clipped to $0.25 \text{m}.\text{s}^{-1}$ in directions $x$ and $y$. 

On the same $500$ randomized environments of Sec. \ref{sec:mcts_results}, this naive contact planner reached the goal $29 \%$ of the times ($82 \%$ for MCTS).
Similarly, for the experiment of Sec. \ref{sec:trotting_gait}, the naive controller had only $7\%$ of success with the trotting gait ($68 \%$ for MCTS).

\subsection{Choice of network  architecture}\label{sec:preliminary_results}

To systematically compare the performance of different neural network architectures, we designed a toy problem. In this problem, the dataset \textit{multimod}, consists of two different contact sequences to reach the same goal (in diagonal) starting from the same position in the same non-randomized environment. One plan goes first straight and left, the other one goes left and then straight to reach the goal. We record 50 randomized runs for each plan for a total of 700 samples in the dataset. Note that the main reason that we performed this experiment on a simple setting is to focus on the multi-modality of the solutions rather than complexity.

Since we perturb the states while collecting data, the dataset is not strictly multi-modal. We would like to evaluate how this partial multi-modality affects the trained policy to reproduce the variety of contact plans of the dataset. Generating diverse contact plans is beneficial as some might be more relevant in specific states.
We compare 3 different model architectures presented in \ref{sec:architecture} on this dataset. The training parameters are detailed in the appendix. 

\subsubsection{Results}

\begin{figure}
  \centering
  \begin{subfigure}{0.32\linewidth}
    \centering
    \includegraphics[width=\linewidth]{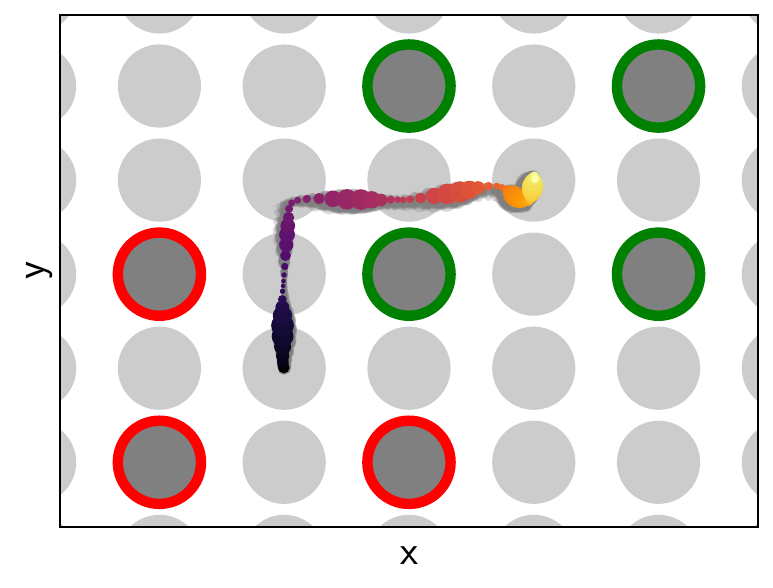}
    \caption{MLP}
    \label{fig:results_diagonalMLP}
  \end{subfigure}
  \begin{subfigure}{0.32\linewidth}
    \centering
    \includegraphics[width=\linewidth]{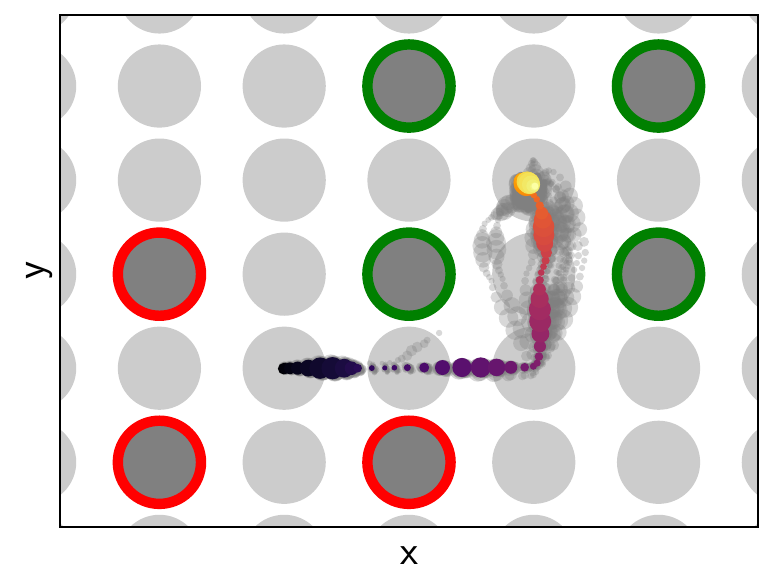}
    \caption{PtrNet}
    \label{fig:results_diagonalPtr}
  \end{subfigure}
  \begin{subfigure}{0.32\linewidth}
    \centering
    \includegraphics[width=\linewidth]{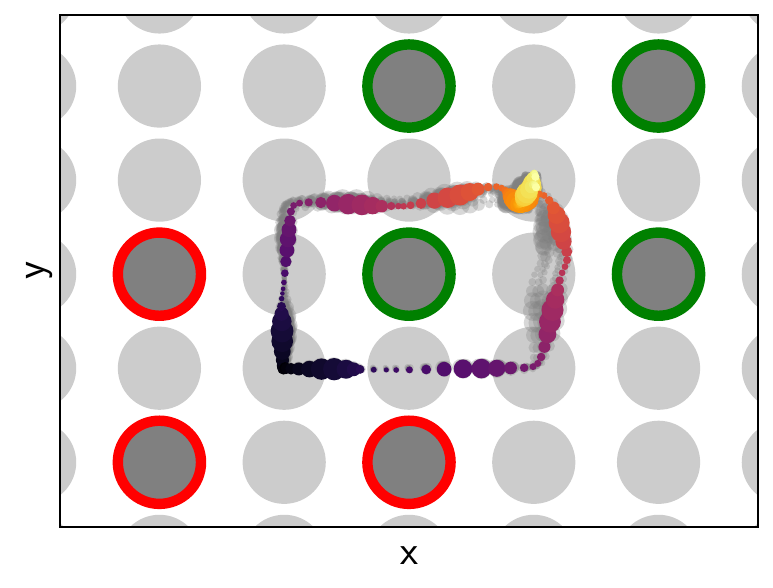}
    \caption{Diffusion}
    \label{fig:results_diagonalDDPM}
  \end{subfigure}
  \caption{Base position $(x,y)$ for 20 randomized runs with models trained on the \textit{multimod} dataset. Starting contact locations are circled in red and goal contact locations in green. The base position is recorded every 20ms. The size of the circles represents the height of the base. Only the diffusion model \ref{fig:results_diagonalDDPM} is able to reproduce both two contact sequences of the training dataset.}
  \label{fig:results_diagonal}
  \vspace{-6mm}
\end{figure}

As shown in Fig. \ref{fig:results_diagonal}, only the diffusion model was able to reproduce the two possible modes of the solution. On 20 simulations, the diffusion model reached the goal by going 12 times first to the right and 8 times first up (we used a different seed each time to generate the initial noise).  MLP and Pointer-Network collapsed to one of the solutions, as can be seen on the base trajectories in Fig.~\ref{fig:results_diagonal}. As we would like to benefit from the multiple feasible solutions of MCTS for a given goal, we focus on diffusion models for the navigation task in the following section.
Note that, for the diffusion model, the projection error $|| \hat{\contact}_B^{i_{j,h}} - p_{\mathcal{C}}(\hat{\contact}_B^{i_{j,h}}) ||_2$  is 5.4mm on average and at a maximum 2.2cm for the predicted jump locations, which is approximately 3 times less than the half distance between two stepping stones.


\subsection{Learning to navigate on stepping stones} \label{sec:navigation_results}
In this subsection, we aim to qualitatively show that the learned policy can reactively generate feasible contact plans when the environment or the goals change on the fly.
To do that, we generated $N_{\text{env}} = 80$ different environments, but this time with less randomization as we focus on the ability of the policy to plan and select the right contact locations: $N_{\text{removed}} = 12$, $\alpha_x = \alpha_y = \alpha_h = 0$. We expect the results to be reproducible in a randomized environment with a larger dataset. The number of goals for each environment $N_{\text{goals}}$ is $8$, $N_{\text{paths}} = 3$ and $N_{\text{repeat}} = 5$. Finding more feasible Our training dataset, \textit{multigoal}, has 52224 samples in total. Our test environment has been generated in the same conditions on $N_{\text{env}} = 20$ different environment. We tested our policy on test goals for which MCTS found a feasible contact plan. Our training procedure and hyperparameters are similar to ~\cite{chi2023diffusion} and are detailed in the appendix.

As done in ~\cite{chi2023diffusion}, we used a DDIM~\cite{song2022denoising} approach to decouple the number of denoising iterations in training and inference. Taking a similar number of steps in the forward and backward diffusion processes usually leads to better results but is time-consuming. For 15 steps in the backward process and above, the policy achieved approximately a similar success rate (our policy was trained on $T = 50$ diffusion steps). Therefore, we used 15 denoising iterations in the experiments. It takes 70 ms for the policy to be evaluated on an AMD Ryzen 5 5625u CPU.

\subsubsection{Static environments}\label{sec:static}


Our policy has been trained to output different contact patches for the next jump as our dataset contains multi-modal samples. Therefore, the final full-length contact plan is likely to differ from the ones of MCTS. This is confirmed as $40\%$ of the successful contact plans were not in the training dataset when replaying on the training environment.
When replayed in the same conditions 20 times for a goal in a corner of the environment, our trained policy reached the goal all the times in 14 different ways. Some examples can be seen in the accompanying \href{https://youtu.be/qvIvTUFh_q4}{video}. This confirms the results of Sec. \ref{sec:preliminary_results} with a policy trained on a more diverse dataset.

\subsubsection{Dynamic environments}\label{sec:dynamic}

Now, we proceed to evaluate how well the policy can be used in a dynamic environment.
To do that,  we randomly remove two contact locations from the environment before each jump while the robot is reaching a goal. The removed stepping stones are chosen among the ones found by MCTS. This way, it is more likely to remove a stepping stone that would have been initially selected, which shows the ability of the policy to replan reactively. In that case, we do not provide the removed locations to the network.
This task is challenging, as removing stepping stones could make the robot jump into a position that is bound to fail or that is out of the training data distribution. Our policy sporadically succeeded on this task ($22\%$ of the time on 50 trials on a goal in diagonal).
Additionally, our trained policy was also able to perform navigation to reach some user-defined changing goals in a new environment with 12 removed stepping stones.
Successful examples of those two tasks can be seen in the accompanying \href{https://youtu.be/qvIvTUFh_q4}{video}. 



\section{Conclusions and future work}\label{sec:conclusion}

In this paper, we presented a framework to efficiently search for non-trivial contact plans in a known environment for legged robots. In our framework, we proposed a customized version of MCTS together with an NMPC to search for feasible solutions on stepping stones. We showed that we can reliably find feasible solutions for different arrangements of the stepping stones as well as initial and final conditions. Collecting feasible rollouts enabled us to collect a rich dataset and learn a control policy that can generate contact plans for the NMPC reactively. Our extensive analyses showed that, due to the multi-modal nature of the dataset, using diffusion models is an ideal way to perform supervised learning on the dataset.

In the future, we plan to explore alternative architectures for diffusion models, as the one used in this study is derived from works with different learning problem structures.
We would like to implement the learned high-level contact planner together with the NMPC on the real hardware. To further reduce the computation time of the whole control pipeline for real-world experiments, we are interested in replacing the NMPC with a learned low-level policy \cite{khadiv2023learning}. We also aim to extend our framework to optimize the gait as well as the surface patches, given a local environment map. Finally, we are interested in learning the high-level policy from partial observation of the robot on-board cameras.

\begin{table*}[t]
    \centering
    \caption{Training parameters for the models. \textit{EMA}: Exponential Moving Average. \textit{wd}: weight decay.}
    \begin{tabular}{c|c|c|c|c|c|c|c}
        \hline
        \toprule
        \textbf{Model} & \textbf{Dataset} & \textbf{Optimizer} & \textbf{LR} & \textbf{LR scheduler} &\textbf{Batch Size} & \textbf{Epochs} & \textbf{Additional Parameters} \\
        \midrule
        \hline
        MLP & \textit{multimod} & Adam & $10^{-3}$ & Exp. decrease (0.996) & 32 & 500 & \makecell{for removed contacts \\ height is set to 0} \\
        Pointer-Network & \textit{multimod} & Adam & $10^{-3}$ & Exp. decrease (0.996) & 8 & 500 & shuffled contact \\
        Diffusion Model & \textit{multimod} & \makecell{AdamW \\ \scriptsize (EMA: 0.75 + wd: $10^{-6}$)}   \normalsize & $10^{-3}$ & \makecell{Cosine annealing \\ \scriptsize w. linear warmup} &  32 & 500 & shuffled contact \\
        Diffusion Model & \textit{multigoal} & \makecell{AdamW \\ \scriptsize (EMA: 0.75 + wd: $10^{-6}$)}   \normalsize & $10^{-3}$ & \makecell{Cosine annealing \\ \scriptsize w. linear warmup} &  64 & 2000 & shuffled contact \\
        \bottomrule
        \hline
    \end{tabular}
    \label{tab:trainingparams}
    \vspace{-5mm}
\end{table*}

\bibliography{master} 
\bibliographystyle{ieeetr}

\section*{Appendix}\label{sec:appendix}
In this appendix, we present the detailed values used for training the networks. Training parameters can be seen on Tab. \ref{tab:trainingparams}.

\subsubsection{MLP}\label{sec:paramMLP}

We used a standard MLP with 4 hidden layers of latent dimension 64 with LeakyReLU activation. The size input dimension is 273 (81 stepping stones locations) and the size output dimension is 24. The total number of trainable parameters of this model is 35736.

\subsubsection{Pointer-Networks}\label{sec:paramPtr}

Both the encoder and decoder are LSTMs with 2 layers and hidden dimension 32. The attention mechanism is also of hidden dimension 32. $[\inputs_{\text{state}}, \inputs_{\text{goal}}]$ is embedded through an MLP with 2 hidden layers of hidden dimensions 16 and 32 with PReLU activation. The total number of trainable parameters is 31521 which is comparable to the MLP architecture considered.

\subsubsection{U-Net1D}\label{sec:paramUNet}

We chose a U-Net with 3 layers with respectively a channel width equal to 64, 128 and 256 and a kernel size of 3 for the convolutions. Convolutions are sliding on the temporal/end-effector dimension. The sinusoidal embedding dimension is 32. The multi-head attention layer has only one head. We used a squared cosine noise schedule with $\beta_1 = 0.004$ and $\beta_T = 0.02$ as suggested by ~\cite{ho2020denoising} for $T = 50$ training iterations.  The model has in total 2639207 parameters.






\end{document}